\documentclass[letterpaper, 10 pt, conference]{ieeeconf}  

\IEEEoverridecommandlockouts

\usepackage[T1]{fontenc}
\usepackage{caption}
\usepackage{cite}

\usepackage{amssymb}
\usepackage{graphicx}
\usepackage{booktabs}
\usepackage{algorithm}
\usepackage{algorithmic}
\usepackage{multirow}
\usepackage{amsmath}
\usepackage{wasysym}
\usepackage{bbm}
\usepackage{mathtools}
\usepackage{tablefootnote}
\usepackage{pifont}
\usepackage[utf8]{inputenc}
\usepackage{float}
\usepackage{placeins}
\usepackage{hyperref}

\title{Towards Exploratory and Focused Manipulation with Bimanual Active Perception: A New Problem, Benchmark and Strategy}

\author{Yuxin He$^{1}$, Ruihao Zhang$^{1 \dagger}$, Tianao Shen$^{1 \dagger}$, Cheng Liu$^{1}$, Qiang Nie$^{1 \star}$
\thanks{$^{1}$The Hong Kong University of Science and Technology (Guangzhou).}
\thanks{$^{\dagger}$Work done during internship.}
}

\begin{document}

\maketitle
\thispagestyle{empty}
\pagestyle{empty}

\begin{abstract}

Recently, active vision has reemerged as an important concept for manipulation, since visual occlusion occurs more frequently when main cameras are mounted on the robot heads. We reflect on the visual occlusion issue and identify its essence as the absence of information useful for task completion. Inspired by this, we come up with the more fundamental problem of \emph{Exploratory and Focused Manipulation} (EFM). The proposed problem is about actively collecting information to complete challenging manipulation tasks that require exploration or focus. As an initial attempt to address this problem, we establish the EFM-10 benchmark that consists of 4 categories of tasks that align with our definition (10 tasks in total). We further come up with a \emph{Bimanual Active Perception} (BAP) strategy, which leverages one arm to provide active vision and another arm to provide force sensing while manipulating. Based on this idea, we collect a dataset named BAPData for the tasks in EFM-10. With the dataset, we successfully verify the effectiveness of the BAP strategy in an imitation learning manner. We hope that the EFM-10 benchmark along with the BAP strategy can become a cornerstone that facilitates future research towards this direction. Project website: \href{EFManipulation.github.io}{EFManipulation.github.io}

\end{abstract}

\section{INTRODUCTION}

With the rapid development of humanoids, the focus of manipulation research is shifting from cobot manipulation for industrial scenarios to humanoid manipulation for general-purpose applications. Along with this shift, an obvious change happens to the placement of the main camera relative to the robotic arm(s). Rather than mount the main camera on the side of the workspace, more researchers now prefer to mount the main camera on the robot's head. This change leads to base-invariant flexibility but has a side effect - the visual occlusion issue (the main view gets occluded more frequently). Some recent studies \cite{11128253, xiong2025via} leverage active vision based on a high-DoF active neck to address it. We notice that the essence of the visual occlusion issue is the absence of information useful for task completion. To perform manipulation tasks with insufficient observations, humans can actively use their senses by, for example, moving their heads or reaching out to touch and feel objects. To endow robots with similar capabilities, we extend the visual occlusion issue to a broader and more fundamental problem beyond it, which is the problem of \textbf{E}xploratory and \textbf{F}ocused \textbf{M}anipulation (\textbf{EFM}).

The core of EFM is \emph{to actively seek information in order to complete challenging manipulation tasks that require exploration or concentration}. For instance, consider a scenario where some audio-video cables of different colors need to be plugged into the corresponding ports that are small and invisible from a fixed view. An agent has to actively explore the color of the ports first in order to pick the correct cable to insert. And the insertion process is delicate, which can benefit from visual information and force information regarding the insertion area. Many extra considerations are required to effectively tackle challenging tasks like this, which is why we find it necessary to formally introduce the problem and call for continuous efforts towards this direction.

As an initial attempt to address the EFM problem, a comprehensive benchmark, EFM-10, is established, where we devise 10 challenging tasks that require exploration or focus. The tasks are grouped into 4 categories based on the different abilities they require:
\begin{itemize}
    \item \textbf{Semantically exploratory tasks}: 1) \emph{Toy-Find}, find a toy of a required color from a random compartment and put it on the table; 2) \emph{Toy-Match}, check the color of a plate in a compartment and put a toy of the same color on the plate.
    \item \textbf{Exploratory tasks involving visual occlusion}: 1) \emph{Cup-Hang}, hang a cup on a rack; 2) \emph{Cup-Place}, place a cup on a small coaster; 3) \emph{Box-Push}, push a box into the lined area.
    \item \textbf{Delicate tasks requiring focus}: 1) \emph{Light-Plug}, plug a USB light; 2) \emph{Bread-Brush}, put a bread dough on the baking tray and brush it; 3) \emph{Nail-Knock}, place a nail and knock it in.
    \item \textbf{Complex tasks requiring both exploration and focus}: 1) \emph{Cable-Match}, check the color of a port and plug in the cable of the same color; 2) \emph{Charger-Plug}, plug in a USB charger.
\end{itemize}

We further come up with a \textbf{B}imanual \textbf{A}ctive \textbf{P}erception (\textbf{BAP}) strategy for bimanual robots to handle EFM tasks. The idea is to \emph{leverage the non-operating arm (if available) to provide  eye-in-hand active vision for the ongoing task and leverage the operating arm to provide force sensing during contact}. This strategy is inspired by the fact that most existing humanoids do not have a 6/7-DoF active neck adopted by recent active vision research \cite{11128253, xiong2025via}, but do have two arms that do not always operate together, which can be utilized in a more efficient way. Note that the BAP strategy is fully compatible with neck-based active vision, and it may be desirable to combine the two, in order to maximize the utilization of all available cameras and to achieve active perception when both arms are busy. We leave this topic for future research.

To verify the effectiveness of the BAP strategy, we collect a dataset called BAPData based on the strategy. Specifically, 1810 expert demonstrations for tasks in EFM-10 are recorded with a real-world bimanual robot. Detailed configurations of the EFM-10 benchmark and the BAPData dataset are introduced in Section \ref{sec:benchmark} and Section \ref{sec:BAP}.

Our experiments demonstrate that the eye-in-hand active vision provided by the non-operating arm can substantially boost performance on all tasks in EFM-10 and the force sensing provided by the operating arm can lead to neural force compliance control that benefit EFM tasks involving fine-grained operation. During the experiments, we also discover a subtle but important technical detail about active vision: It is better to capture the operating end effector in the active view when delicate operation is being conducted with a hand-held object, as capturing the hand-held object alone in the active view cannot provide \emph{direct} clues about how the operating end effector should adjust its pose.

To sum up, our main contributions include:
\begin{itemize}
    \item We come up with the problem of Exploratory and Focused Manipulation (EFM) and construct a comprehensive benchmark (EFM-10) as an initial step to address it.
    \item We introduce the Bimanual Active Perception (BAP) strategy that helps address EFM tasks without requiring a high-DoF active neck and collect the BAPData dataset for tasks in EFM-10 based on this strategy.
    \item We verify the effectiveness of BAP and benchmark the performance of representative policy models in EFM-10 by conducting imitation learning with BAPData.
\end{itemize}

\section{RELATED WORK}

\subsection{Active Vision for Manipulation}

Active vision has been studied for 3D reconstruction \cite{krainin2011autonomous}, object detection \cite{le2008active}, object tracking \cite{liu2020target} and other applications related to view planning \cite{zeng2020view}. Active vision also has been introduced for manipulation by \cite{van2021active}, where a single robotic arm is responsible for manipulating as well as observing with its in-hand camera. More recently, AV-ALOHA \cite{11128253} has proposed leveraging a 7-DoF active neck to achieve active vision for humanoid manipulation, in order to deal with the more frequently-occurring visual occlusion. ViA \cite{xiong2025via} adapts a similar idea as AV-ALOHA and proposes some technical improvements in terms of the teleoperation system. In contrast, we propose to leverage the non-operating arm (if available) to provide an active view for manipulation, which is more readily applicable to existing humanoids.

\subsection{Active Exploration for Manipulation}

Active exploration is a relatively mature concept in the field of navigation research, but under-explored for manipulation. Existing work related to active exploration for manipulation focuses mainly on identifying the system dynamics in the real world through active exploration to facilitate model-based control \cite{9982061,memmelasid}. More recently, RoboEXP \cite{jiang2025roboexp}, a robotic exploration system based on large multimodal models, has been proposed for interactive scene exploration. However, RoboEXP focuses on scene graph construction and the application of scene graph on manipulation, which is less relevant to learning-based manipulation. EyeRobot \cite{kerr2025eye} is the first work that introduces the concept of active exploration for end-to-end manipulation policy learning. And this paper is the first work that explores this for bimanual settings.

\subsection{Force Sensing for Manipulation}

Force sensing is crucial for providing hardly visible contact information required by fine-grained manipulation. There are mainly two kinds of force sensing: 6D Force/Torque and Visuo-Tactile. 6D Force/Torque has been widely applied to classical compliance control \cite{4308708}. Some studies also employ 6D Force/Torque for learning-based contact-rich manipulation via reinforcement learning \cite{6095096} or self-supervised representation learning \cite{6095096}. Visuo-Tactile \cite{5206534} mimics human haptics. Recently, many works \cite{huang20253d, ding2024bunny, li2024haptic, xue2025reactive, He2025ViTacFormerLC} combine Visuo-Tactile with imitation learning to address fine-grained manipulation tasks that involve hand-held objects. Our work employs 6D Force/Torque for end-to-end learning of fine-grained manipulation, since Force/Torque sensors are more commonly available in existing robots.

\begin{figure*}
\centering
\includegraphics[width=0.89\linewidth]{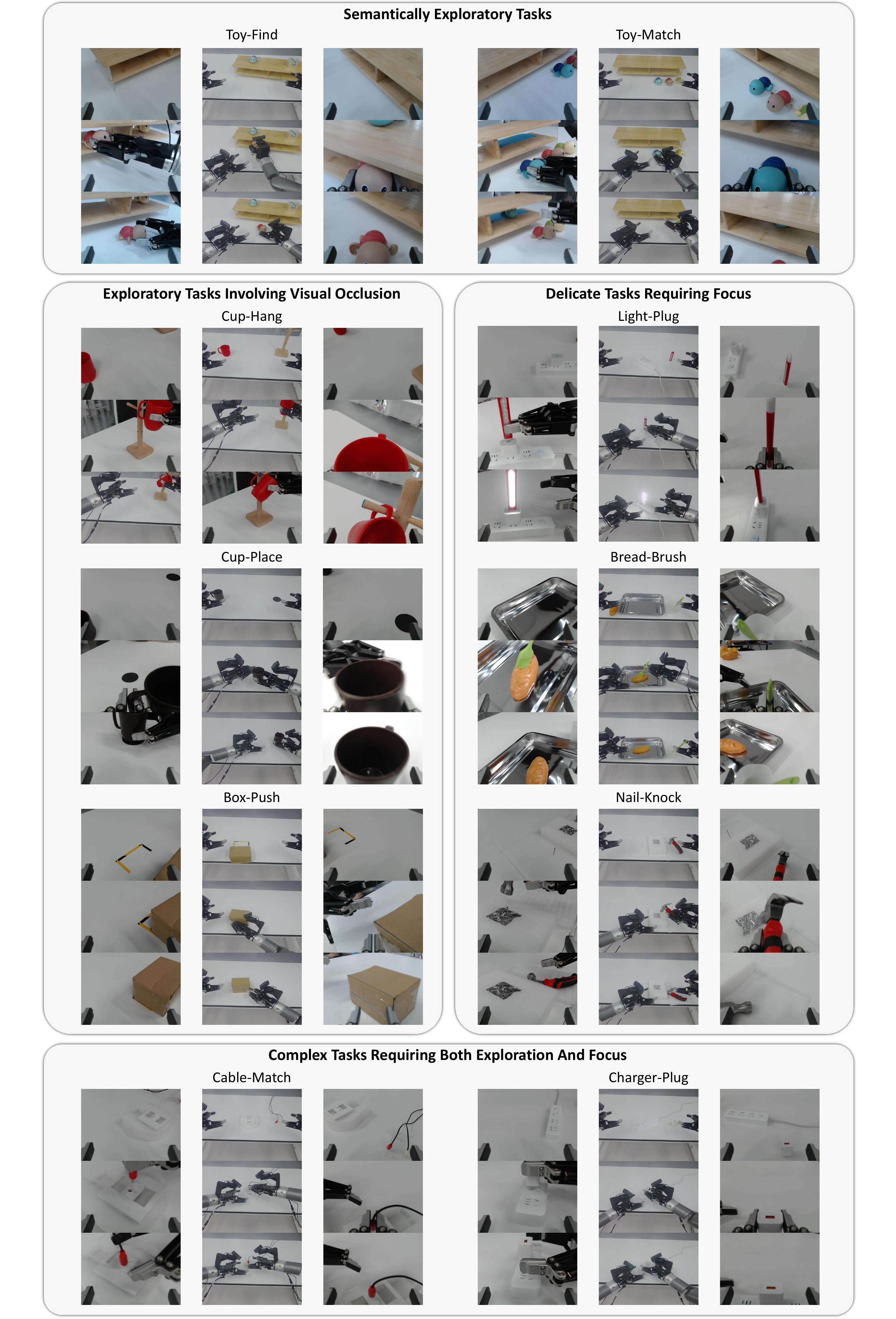}
\caption{An illustration of the BAPData dataset collected for the 10 tasks in EFM-10. The left wrist-view images (left column), main-view images (middle column), and right wrist-view images (right column) are shown temporally for each example.}
\label{fig:data}
\end{figure*}

\begin{table*}[t]
    \centering
    \caption{Comparison of Real-world Humanoid Manipulation Datasets}
    \label{datasets}
    \scalebox{1.0}[1.0]{
	\begin{tabular}{l|cccccc}
		\toprule
        \multirow{2}{*}{Dataset} & \multicolumn{2}{c}{Active Vision} & \multirow{2}{*}{6/7-DoF Active Neck} & \multirow{2}{*}{Force} & \multirow{2}{*}{\# Tasks} & \multirow{2}{*}{\# Traj.} \\
        \cmidrule(lr){2-3} & Low-DoF & High-DoF & & & & \\
        \midrule
        ALOHA \cite{zhao2023learning}  &  &  &  &  & 6 & 350 \\
        BRMData \cite{zhang2024empowering} & & & & & 10 & 500 \\
        Galaxea Open-World \cite{jiang2025galaxea} &  &  &  &  & 150 & 100K \\
        AgiBot World \cite{bu2025agibot} &  &  &  & \checkmark & 217 & 1M+ \\
        Open-Television \cite{Cheng2024OpenTeleVisionTW} & \checkmark &  &  &  & 4 & 80 \\
        ViTacFormer \cite{He2025ViTacFormerLC} & \checkmark &  &  & \checkmark & 5 & 300 \\
        AV-ALOHA \cite{11128253} & & \checkmark & Required &  & 7 & 350 \\
        ViA \cite{xiong2025via} & & \checkmark & Required &  & 3 & 535 \\
        \midrule
        BAPData (Ours) & & \checkmark & & \checkmark & 10 & 1850 \\
		\bottomrule
	\end{tabular}
    }
\end{table*}

\section{The EFM-10 Benchmark}
\label{sec:benchmark}

The EFM-10 benchmark includes 4 categories of tasks that require different capabilities. We define each category as follows:
\begin{itemize}
    \item \textbf{Semantically exploratory tasks} refer to tasks in which a robot has to actively explore the \emph{hidden} semantic properties of a scene in order to fulfill the semantic requirement with the task command.
    \item \textbf{Exploratory tasks involving visual occlusion} refer to tasks in which a robot will fail to capture what is happening around the manipulated area if the robot does not have an active view.
    \item \textbf{Delicate tasks requiring focus} refer to tasks in which a robot has to carry out fine-grained operation that can greatly benefit from a concentrated (and clearer) view of the manipulated area.
    \item \textbf{Complex tasks requiring both exploration and focus} refer to tasks in which a robot will have to explore the hidden semantic properties of a scene or avoid visual occlusion with an active view, and carry out fine-grained operation with the help of a concentrated view.
\end{itemize}

Detailed specifications of each category of tasks within EFM-10 are provided in the following subsections. We try to standardize the task definitions so that researchers around the world can instantiate the tasks in simulation or in the real world with their own robots.

\subsection{Semantically Exploratory Tasks}

\subsubsection{Toy-Find}
There is a cabinet with at least two compartments. A toy of the required color is placed in one of the compartments, and there may be toys of other colors in all the compartments. All toys are initially invisible to the robot. The task instruction is ``Pick a \emph{[The-Required-Color]} toy from one of the compartments of the cabinet and place it on the table''. To achieve this, the robot is supposed to first check the compartments to find out where the target toy is, then pick the toy and place it on the table.

\subsubsection{Toy-Match}
There are at least two toys of different colors and a cabinet with at least one compartment. In a specified compartment, there is a plate of the same color as one of the toys. The plate is initially invisible to the robot. The task instruction is ``Pick the toy of the same color as the plate in \emph{[The-Specified-Compartment]} and place it on the plate''. To achieve this, the robot is supposed to first check the color of the plate, then pick the toy of the same color from the table and place it on the plate.

\subsection{Exploratory Tasks Involving Visual Occlusion}

\subsubsection{Cup-Hang}

There is a cup rack on one side of a table and at least one cup on the other side of the table. Only one cup is of the required color. The task instruction is ``Pass the \emph{[The-Required-Color]} cup and hang it onto the rack''. During the hanging operation, the hand-held cup will cause visual occlusion, which requires the robot to observe with an active view.

\subsubsection{Cup-Place}

The Cup-Place is similar to the Cup-Hang task. The differences are that the rack is replaced by a coaster of the same size as the cup, and the hanging operation is replaced by the placing operation.

\subsubsection{Box-Push}

There is a box and a lined area on the table. The task instruction is ``Push the box to the lined area''. During pushing, the box will cause visual occlusion, which requires the robot to observe with an active view.

\subsection{Delicate Tasks Requiring Focus}

\subsubsection{Light-Plug}

There is a USB-A or USB-C light and a charger. The task instruction is ``Plug the USB light into the charger''. Since the USB port is small, the robot will benefit from an active view concentrated on the area around the port.

\subsubsection{Bread-Brush}

There is a bread dough, a baking tray, and a brush sitting in a cup of oil. The task instruction is ``Place the bread dough on the tray and brush it with oil''. To better control the contact between the brush and the dough during brushing, it is desirable to have an active view concentrated on the contact area.

\subsubsection{Nail-Knock}

There is a nail, a hammer, and a scrap of silver paper lying on a block. The task instruction is ``Place the nail on the scrap of silver paper and knock the nail in''. Knocking a nail requires precise perception of the relative position between the nail and the hammer face, which will be difficult relying purely on a fixed view.

\subsection{Complex tasks requiring both exploration and focus}

\subsubsection{Cable-Match}
There are at least two cables of different colors and a port of the same color as one of the cables. The color of the port is initially invisible to the robot. The task instruction is ``Insert the cable of the same color as the port''. To achieve this, the robot is supposed to first check the color of the port, then take the corresponding cable and plug the cable into the port.

\subsubsection{Charger-Plug}
There is a USB charger and a power strip with 2\textasciitilde3 ports. The task instruction is ``Plug the USB charger into the \emph{[Left/Middle/Right]} port of the power strip''. During plugging, the charger will cause visual occlusion, hence the robot has to actively find a non-occluded angle of view and concentrate the view on the insertion area when conducting the delicate operation.

\begin{table}[t]
    \centering
    \caption{Task-wise statistics of BAPData. ``\# Variations'' means the number of the combinations of target objects, and object placements in the scene (the variations of distractors are not counted here). For instance, in the data of Toy-Find task, there are 4 possible target toys of different colors and 2 possible compartments where the target toy can be found, so \# Variations is 8.}
    \label{statistics}
    \scalebox{1.0}[1.0]{
	\begin{tabular}{l|cccc}
		\toprule
        Task & \# Variations & \# Traj. & Avg. Len. (Seconds) \\
        \midrule
        Toy-Find & 8 & 160 & 14.6 \\
        Toy-Match & 5 & 140 & 15.1 \\
        Cup-Hang & 9 & 180 & 17.8 \\
        Cup-Place & 6 & 120 & 13.3 \\
        Box-Push & 4 & 150 & 24.5 \\
        Light-Plug & 6 & 300 & 14.7 \\
        Bread-Brush & 20 & 200 & 24.4 \\
        Nail-Knock & 1 & 150 & 19.2 \\
        Cable-Match & 8 & 150 & 15.6 \\
        Charger-Plug & 12 & 300 & 15.1 \\
		\bottomrule
	\end{tabular}
    }
\end{table}

\begin{figure*}
\centering
\includegraphics[width=0.95\linewidth]{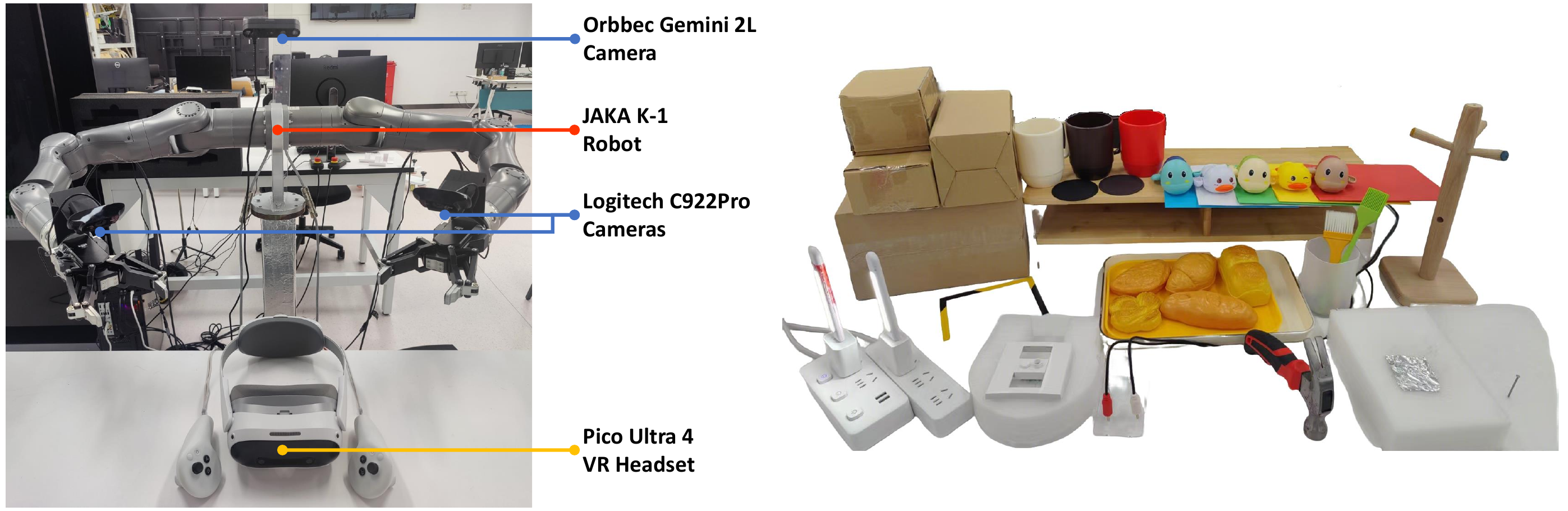}
\caption{An overview of the hardware system and the objects.}
\label{robot_objects}
\end{figure*}

\begin{table*}[ht]
    \centering
    \caption{Comparison of performances under different settings of visual context captured by the active view. Succuss rate (\%) of each task under each setting is reported here.}
    \label{verify_ehav}
    \scalebox{1.0}[1.0]{
	\begin{tabular}{cccccc}
		\toprule
        \multicolumn{2}{c}{Visibility in the Active View} & \multirow{2}{*}{Toy-Match} & \multirow{2}{*}{Cup-Hang} & \multirow{2}{*}{Nail-Knock} & \multirow{2}{*}{Charger-Plug} \\
        \cmidrule(lr){1-2} Manipulated  area & End Effector & & & \\
        \midrule
        & & 20.0 & 23.3 & 6.67 & 0.00 \\
        \checkmark & & - & 76.7 & 26.7 & 13.3 \\
        \checkmark & \checkmark & 76.7 & 90.0 & 43.3 & 20.0 \\
		\bottomrule
	\end{tabular}
    }
    \vspace{-0.3em}
\end{table*}

\section{Bimanual Active Perception}
\label{sec:BAP}

There are many possible strategies that may address (some of) EFM tasks. For instance, to complete a semantically exploratory task, a robotic arm can first explore the environment and then operate \cite{jiang2025roboexp}; to complete an exploratory task involving visual occlusion, a high-DoF active neck can be employed to provide an active view \cite{11128253, xiong2025via}. In this work, we introduce a new strategy that is readily applicable to existing bimanual robots without requiring an active neck. Specifically, we propose to leverage the non-operating arm (if available) to provide high-DoF eye-in-hand active vision and leverage the operating arm to provide force sensing during contact. There are many possible ways to implement such a strategy, such as imitation learning and reinforcement learning. And we choose imitation learning as it is so far the most effective paradigm for general-purpose manipulation.

Hence, a high-quality imitation learning dataset, BAPData, is constructed based on the BAP strategy. BAPData consists of 1850 expert demonstrations collected for the 10 tasks in EFM-10 with a real-world bimanual robot. See Fig. \ref{fig:data} for some example of the data. During teleoperation, the expert demonstrator intentionally controls the available non-operating arm to provide an active view that captures both the manipulated area and the operating end effector when exploration or focus is needed. Force/Torque sensory data are recorded in the dataset as part of the bimanual active perception to facilitate delicate manipulation tasks that involve fine-grained contact. Table \ref{datasets} compares the BAPData dataset with existing real-world bimanual manipulation datasets. It shows that our dataset is so far the biggest one with active vision, and the only one with both high-DoF active vision and force information.

Our hardware system and the objects that we used for data collection are shown in Fig. \ref{robot_objects}. The hardware system consists of a JAKA K-1 bimanual robot, an Orbbec Gemini 2L camera fixed on the head, two Logitech C922Pro cameras mounted on the wrists, and a Pico Ultra 4 for VR-based teleoperation. The JAKA K-1 robot has two 7-DoF arms, each of which has a built-in Force/Torque sensor.

All data are recorded with a frequency of 10 Hz. Statistics of the data collected for each task are listed in Table \ref{statistics}.

\section{EXPERIMENTS}

\subsection{Verifying Eye-in-Hand Active Vision}

Preliminary experiments have been conducted to verify the idea of leveraging the non-operating arm to provide eye-in-hand active vision for EFM. In the preliminary experiments, we consider two additional settings of visual context captured by the active view. In the first setting, neither the manipulated area nor the operating end effector is captured. In the second setting, only the manipulated area is captured and the operating end effector is out of the active view (this only happens when the end effector is manipulating a fair-sized hand-held object or tool).

We choose 4 tasks from the 4 categories of the EFM-10 benchmark and, for each additional setting, collect the same amount of data for these tasks as in BAPData. We then train ACT \cite{zhao2023learning} policies for each of the 4 tasks under the 3 different settings of visual context in an imitation learning manner. The model inputs include images from the head, left-wrist, right-wrist cameras and the robot state (the poses of left, right end effectors and left, right gripper states, dimension=14+2). The model output is an action chunk of size 8 within the Cartesian action space. After that, we evaluate each policy in 30 random trials.

\begin{table*}[t]
    \centering
    \caption{Evaluation results of representative manipulation policy models on our EFM-10 benchmark. ``$\star$'' means that we train the policy with BAPData collected based on our BAP strategy. Force sensing is not included in this experiment.}
    \label{benckmarking}
    \scalebox{0.92}[0.92]{
	\begin{tabular}{l|cccccccccc}
		\toprule
         \multirow{2}{*}{Method} & \multicolumn{2}{c}{Semantically Exploratory} & \multicolumn{3}{c}{Exploratory (Visual Occlusion)} & \multicolumn{3}{c}{Delicate (Requiring Focus)} &
         \multicolumn{2}{c}{Exploratory \& Focused} \\
         \cmidrule(lr){2-3} \cmidrule(lr){4-6} \cmidrule(lr){7-9} 
         \cmidrule(lr){10-11} &
         Toy-Find & Toy-Match & Cup-Hang & Cup-Place & Box-Push & Light-Plug & Bread-Brush & Nail-Knock & Cable-Match & Charger-Plug \\
        \midrule
        ACT\cite{zhao2023learning} & - & 20 & 23.3 & - & - & - & - & 6.67 & - & 0.0 \\
        ACT\cite{zhao2023learning}$^\star$ & 26.7 & 76.7 & 90.0 & 93.3 & 40.0 & 23.3 & 80.0 & 43.3 & 66.7 & 20.0 \\
        DP\cite{chi2023diffusion}$^\star$ & 30.0 & 83.3 & 93.3 & 86.7 & 53.3 & 20.0 & 80.0 & 36.7 & 53.3 & 13.3 \\
        \midrule
        GR-MG\cite{li2024gr}$^\star$ & 60.0 & 86.7 & 93.3 & 93.3 & 46.7 & 20.0 & 76.7 & 46.7 & 70.0 & 23.3 \\
        Pi-0\cite{black2024pi}$^\star$ & 83.3 & 80.0 & 96.7 & 100 & 63.3 & 23.3 & 83.3 & 40.0 & 63.3 & 16.7   \\
		\bottomrule
	\end{tabular}
    }
\end{table*}

\begin{figure}[t]
\centering
\includegraphics[width=0.9\columnwidth]{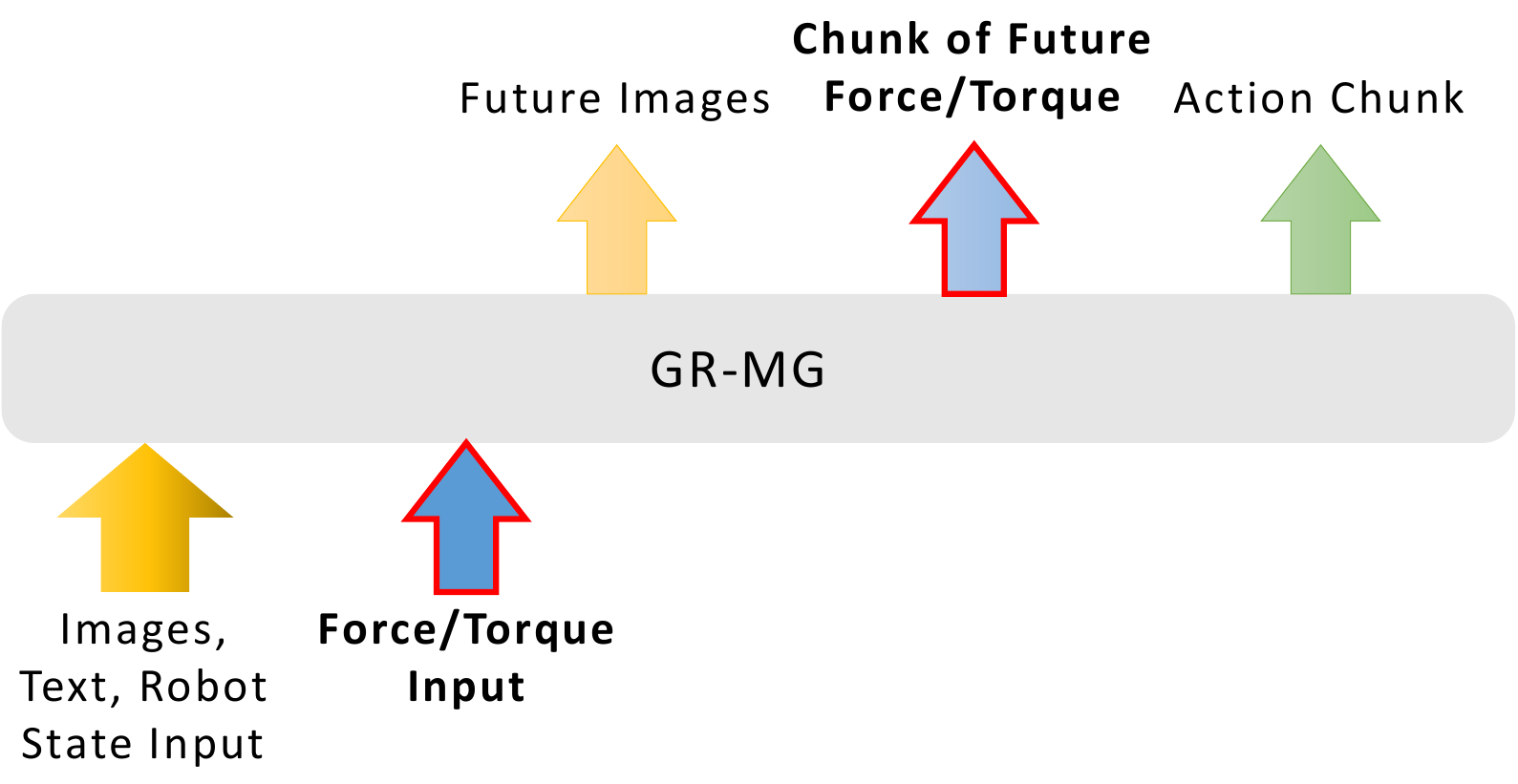}
\caption{A way to incorporate force sensing into the GR-MG policy, where we include Force/Torque as extra input and train the model to additionally predict future Force/Torque.}
\label{fig:model}
\vspace{-0.5em}
\end{figure}

The experiment results are shown in Table \ref{verify_ehav}. We can observe that: 1) If the active view cannot capture both the manipulated area and the operating end effector, the policy tends to fail the EFM tasks. 2) If the active view only captures the manipulated area but not the operating end effector, the performance is inferior to the case where both the manipulated area and the operating end effector are captured by the active view. This may be due to the random and diverse relative pose between the hand-held object and the end effector, which prevents direct judgment about how the end effector should adjust its pose relying purely on the pose of the hand-held object. These results not only support the idea of leveraging the non-operating arm to provide eye-in-hand active vision but also point out a subtle while important technical detail when implementing active vision, that is: It is desirable to capture the operating end effector in the active view.

\subsection{Benchmarking Representative Policies}

Extensive evaluation of the performance of representative manipulation policies on the proposed EFM-10 benchmark is carried out to find out the pros and cons of these policies. The following polices are considered:
\begin{itemize}
    \item ACT \cite{zhao2023learning}, a single-task transformer-based policy that leverages action chunking.
    \item DP \cite{chi2023diffusion}, a single-task policy based on the diffusion model. (We use the CNN version in our experiments.)
    \item GR-MG \cite{li2024gr}, a multi-task transformer-based policy that leverages future image prediction \cite{wu2023unleashing,3DForesight} and action chunking.
    \item Pi-0 \cite{black2024pi}, a multi-task policy that combines a VLM and flow matching.
\end{itemize}

We use BAPData to train policy models in a imitation learning manner (without using Force/Torque sensory data). The
state/action space for all the policies is the Cartesian state/action space. For each single-task algorithm, we train a separate model for each task respectively; for each multi-task algorithm, we train a model for all tasks.  Table \ref{benckmarking} shows the overall evaluation results. It can be observed that:
\begin{itemize}
    \item Single-task policies ACT and DP cannot fulfill language-driven semantically exploratory tasks like Toy-Find, as they are not conditioned on language.
    \item DP is superior to ACT on handling tasks with multimodal action distributions (such as Box-Push), but inferior to ACT on handling fine-grained tasks like Nail-Knock and Charger Plug (possibly because DP does not employ temporal ensemble).
    \item Pi-0 show stronger instruction-following ability than GR-MG and can more easily master EFM tasks that do not involve fine-grained operation, such as Toy-Find, Cup-Hang and Cup-Place.
    \item All these policies do not perform well EFM tasks that involve extremely fine-grained operation, such as Light-Plug and Charger-Plug.
\end{itemize}

\begin{table}[t]
    \centering
    \caption{Comparison of the performance of GR-MG with and without force sensing. ``Avg. Fz Max'' means the average of maximum vertical force of the operating end-effector. ``SR'' means task success rate (\%).}
    \label{verify_force}
    \scalebox{0.94}[0.94]{
	\begin{tabular}{l|cccc}
		\toprule
         \multirow{2}{*}{Method} & \multicolumn{2}{c}{Light-Plug} & \multicolumn{2}{c}{Bread-Brush} \\
         \cmidrule(lr){2-3} \cmidrule(lr){4-5} & Avg. Fz Max & SR & Avg. Fz Max & SR \\
        \midrule
        GR-MG & 3.5 & 20.0 & -16.4 & 76.7 \\
        GR-MG + Force/Torque & -0.2 ($\downarrow29\%$) & 36.7 & -17.9 ($\downarrow22\%$) & 90.0 \\
		\bottomrule
	\end{tabular}
    }
    \vspace{-0.3em}
\end{table}

\subsection{The Effect of Force Sensing}

We further carry out experiments on enhancing policies with the force sensing provided by the operating arm. Concretely, a variant of the GR-MG policy is devised, as illustrated in Fig. \ref{fig:model}. We encode current Force/Torque value with a linear layer, append the encoded embedding as well as a query token to the original input sequence, and train the model to predict the chunk of future Force/Torque based on the final representation of the query token.

\begin{figure*}[t]
\centering
\includegraphics[width=1.0\linewidth]{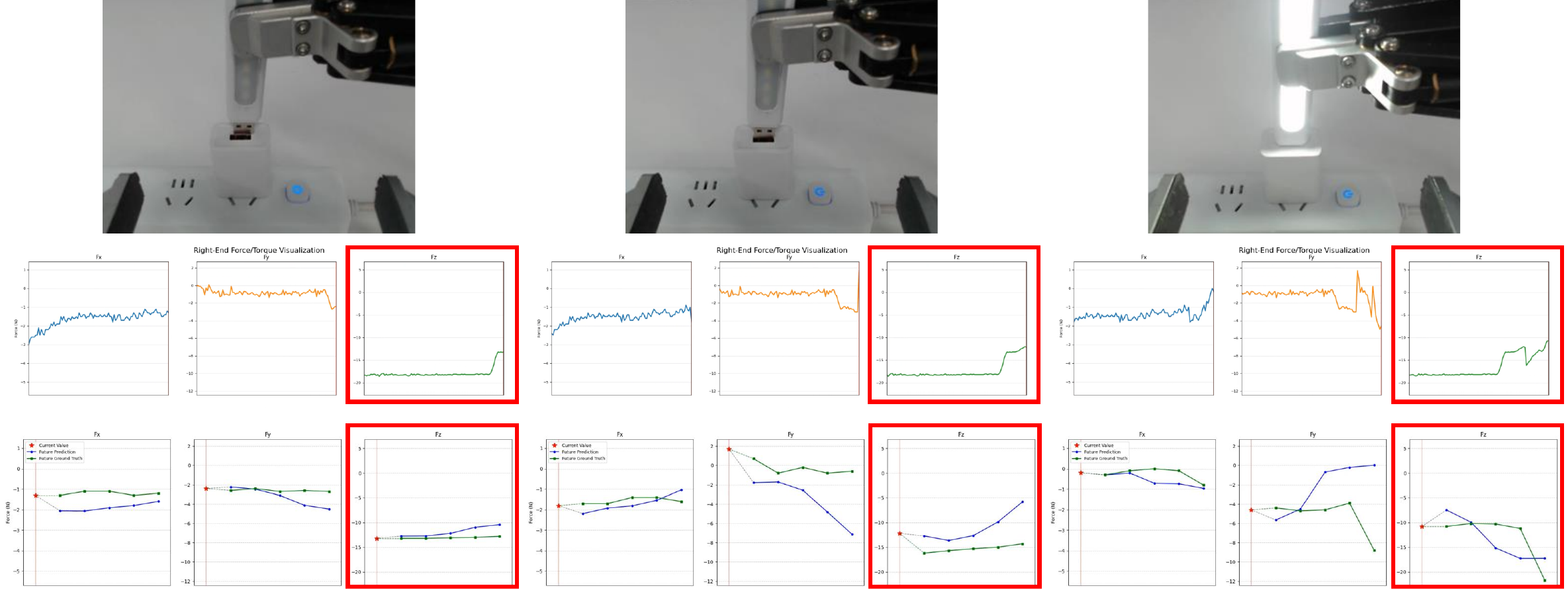}
\caption{Visualization of a rollout by the GR-MG policy with force sensing. In the first row, the images captured by the active view are shown. In the second row, real-time $x, y, z$ Force values of the operating end effector are visualized. In the third row, the predicted future $x, y, z$ Force values are visualized.}
\label{fig:force}
\vspace{-0.5em}
\end{figure*}

Two EFM tasks that involve fine-grained operation are considered here, including Light-Plug and Bread-Brush. The experiment results are shown in Table \ref{fig:force}. With force sensing, the success rates increase by 16.7 and 13.3 on the two tasks. More importantly, the averages of maximum vertical force of the operating end-effector reduce by 29$\%$ and 22$\%$ (relative to the average ranges of vertical force) on the tasks, suggesting that a sort of force compliance control is achieved with our neural network.

To qualitatively analyze this phenomenon, we visualize a rollout by the GR-MG policy with force sensing in Fig. \ref{fig:force}. It can be observed that: When the hand-held USB light began to contact with one side of the USB port, the model forecasted that the vertical force would increase and was able to control the end effector in a way that prevents abrupt boost of vertical force. When the USB light got plugged into the port, the model also correctly forecasted that the vertical force would decrease. These observations strongly support the significance of force sensing.

\subsection{Analysis of Failure Cases}

\begin{figure*}[t]
\centering
\includegraphics[width=1.0\linewidth]{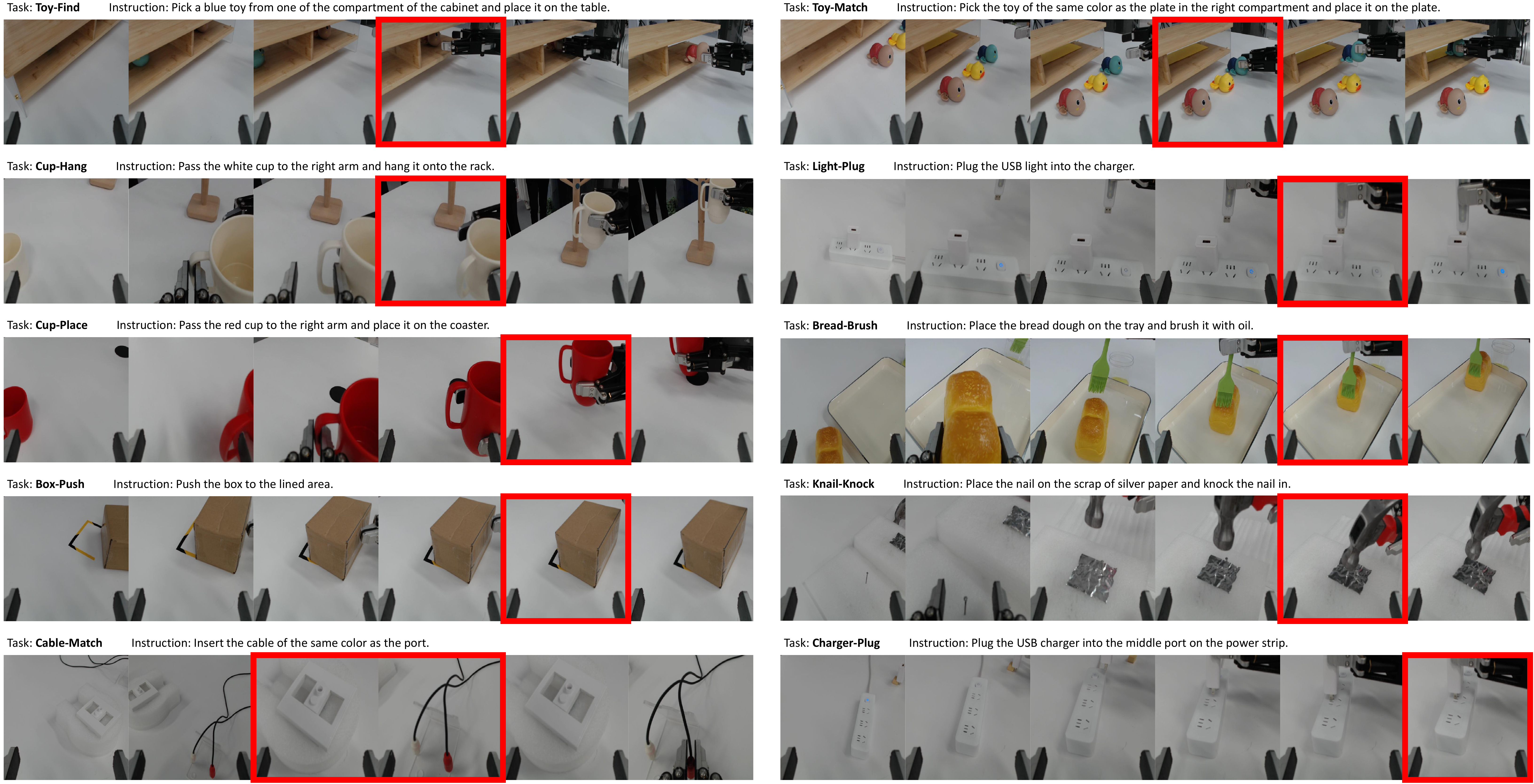}
\caption{Typical failure cases by the policies that we trained with BAPData using imitation learning.}
\label{fig:cases}
\vspace{-0.5em}
\end{figure*}

To provide more insights about how to further improve manipulation policies in order to address EFM tasks, we qualitatively analyze the typical failure modes that we observed during our experiments, as shown in Fig. \ref{fig:cases}. For tasks involving semantical exploration (e.g. Toy-Find, Toy-Match and Cable-Match), most failures are caused by unable to accurately condition the action on the semantic context, e.g., picking the toy or cable of a wrong color. For tasks involving visual occlusion (e.g. Cup-Hang, Cup-Place and Box-Push), the failure modes are diverse. In the Cup-Hang task, many failures are caused by the cup is held overly low by the right gripper after transfer and the right arm fails to adapt to the height. In the Cup-Place task, many failures are caused by the policies fail to find an optimal angle of active view to avoid the sight of coaster from being blocked by the cup. In the Box-Push task, many failures are caused by the policies fail to adapt to the situation when the box oversteps the left boundary. For tasks involving delicate operation (e.g. Light-Plug, Bread-Brush, Knail-Knock and charger-Plug), many failures are caused by the inadequate spatial perception/reasoning capability of the policies, which frequently leads to subtle wrong positioning. Future work should make efforts to improve the aforementioned abilities of policy models.

\section{CONCLUSION}

In this paper, we investigate a novel problem: Exploratory and Focused Manipulation, with the ultimate goal of endowing robots with capabilities similar to humans in active exploration and concentration. EFM-10, a comprehensive benchmark with 4 categories of tasks, is established for this problem. We also propose the Bimanual Active Perception strategy that addresses EFM without requiring a high-DoF active neck. BAPData, a high-quality dataset with 1810 expert demonstrations are collected based on the BAP strategy to facilitate learning-based implementation of the strategy. Our experiments with EFM-10 and BAPData demonstrate the effectiveness of BAP and reveal the pros and cons of representative policies. In-depth analysis of failure cases further suggests some directions for future research on EFM, including enhancing the following capabilities of policy models: semantical conditioning, spatial perception/reasoning, and optimal active viewpoint searching.

\section{ACKNOWLEDGMENT}

This work is supported by the Guangdong Basic and Applied Basic Research Foundation (No. 2026A1515012291).

\bibliographystyle{IEEEtran}
\bibliography{references}


\end{document}